\documentclass[10pt,twocolumn,letterpaper]{article}

\usepackage{iccv}
\usepackage{times}
\usepackage{epsfig}
\usepackage{graphicx}
\usepackage{amsmath}
\usepackage{amssymb}
\usepackage[font=small,labelfont=bf]{caption}

\usepackage[ruled]{algorithm2e}
\usepackage{multirow}
\usepackage{array}
\usepackage{ctable} 
\usepackage{mathptmx}
\usepackage{makecell}
\usepackage{arydshln}
\usepackage[capposition=bottom]{floatrow}
\usepackage{comment}
\DeclareMathAlphabet{\mathcal}{OMS}{cmsy}{m}{n}
\usepackage{textcomp}
\usepackage{dblfloatfix}
\usepackage{caption}
\usepackage{subcaption}

\newcolumntype{?}[1]{!{\vrule width #1}}

\usepackage[pagebackref=true,breaklinks=true,letterpaper=true,colorlinks=false,bookmarks=false]{hyperref}

\iccvfinalcopy 

\def\iccvPaperID{1597} 

\ificcvfinal\fi
\begin{document}

\title{Co-Separating Sounds of Visual Objects}
\author{Ruohan Gao\\
UT Austin\\
{\tt\small rhgao@cs.utexas.edu}
\and
Kristen Grauman\\
UT Austin and Facebook AI Research\\
{\tt\small grauman@cs.utexas.edu}
}

\maketitle


\begin{abstract} 
Learning how objects sound from video is challenging, since they often heavily overlap in a single audio channel. Current methods for visually-guided audio source separation sidestep the issue by training with artificially mixed video clips, but this puts unwieldy restrictions on training data collection and may even prevent learning the properties of ``true" mixed sounds. We introduce a \emph{co-separation} training paradigm that permits learning object-level sounds from unlabeled multi-source videos. Our novel training objective requires that the deep neural network's separated audio for similar-looking objects be consistently identifiable, while simultaneously reproducing accurate video-level audio tracks for each source training pair. Our approach disentangles sounds in realistic test videos, even in cases where an object was not observed individually during training. We obtain state-of-the-art results on visually-guided audio source separation and audio denoising for the MUSIC, AudioSet, and AV-Bench datasets. 

\end{abstract}

\section{Introduction}
\label{sec:intro}

Multi-modal perception is important to capture the richness of real-world sensory data for objects, scenes, and events. The sounds made by objects, whether actively generated or incidentally emitted, offer valuable signals about their physical properties and spatial locations---the cymbals crash on stage, the bird tweets up in the tree, the truck revs down the block, the silverware clinks in the drawer.

Objects often generate sounds while coexisting or interacting with other surrounding objects. Thus, rather than observe them in isolation, we hear them intertwined with sounds coming from other sources. Likewise, a realistic video records the various objects with a single audio channel that mixes all their acoustic frequencies together. Automatically \emph{separating} the sounds of each object in a video is of great practical interest, with applications including audio denoising, audio-visual video indexing, instrument equalization, audio event remixing, and dialog following.

\begin{figure}
	\centering
	\includegraphics[scale=0.54]{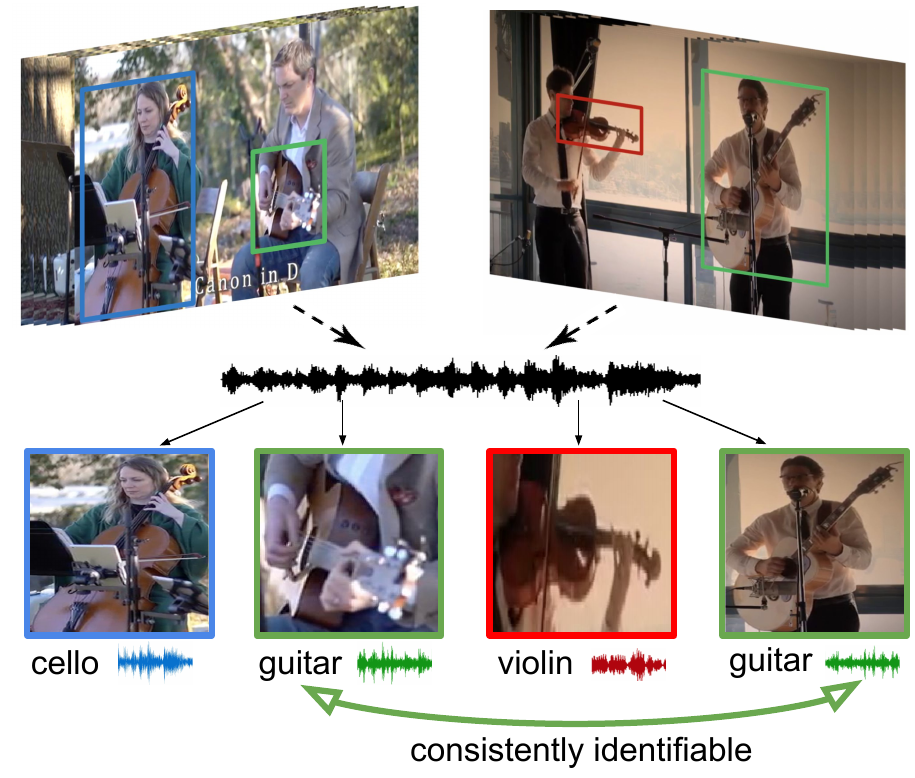}
	\caption{We propose a \emph{co-separation} training objective to learn audio-source separation from unlabeled video containing multiple sound sources.  Our approach learns to associate consistent sounds to similar-looking objects across pairs of training videos.  Then, given a single novel video, it returns a separate sound track for each object. Image credit:~\cite{video1,video2}.}
	\label{fig:concept}
	\vspace*{-0.15in}
\end{figure}
 
Whereas traditional methods assume access to multiple microphones or carefully supervised clean audio samples
~\cite{hyvarinen2000independent,virtanen2007monaural,fevotte2009nonnegative}, recent methods tackle the audio(-visual) source separation problem using a ``mix-and-separate" paradigm to train deep neural networks in a self-supervised manner~\cite{simpson2015deep,yu2017permutation,ephrat2018looking,owens2018audio,zhao2018sound}. Namely, such methods randomly mix audio/video clips, and the learning objective is to recover the original unmixed signals. For example, one can create ``synthetic cocktail parties" that mix clean speech with other sounds~\cite{ephrat2018looking}, add pseudo ``off-screen" human speakers to other real videos~\cite{owens2018audio}, or superimpose audio from clips of musical instruments~\cite{zhao2018sound}.

There are two key limitations with this current training strategy. First, it implicitly assumes that the original real training videos are dominated by single-source clips containing one primary sound maker. However, gathering a large number of such clean ``solo" recordings is impractical and will be difficult to scale beyond particular classes like human speakers and musical instruments. Second, it implicitly assumes that the sources in a recording are independent. However, it is precisely the correlations \emph{between real sound sources} (objects) that make the source separation problem most challenging at test time.  Such correlations can go uncaptured by the artificially mixed training clips.

Towards addressing these shortcomings, we introduce a new strategy for learning to separate audio sources.  Our key insight is a novel \emph{co-separation} training objective that learns from naturally occurring multi-source videos.\footnote{Throughout, we use ``multi-source video" as shorthand for video containing multiple sounds in its single-channel audio.}
During training, our co-separation network considers pairs of training videos and, rather than simply separate their artificially mixed soundtracks, it must also generate audio tracks that are \emph{consistently identifiable at the object level} across all training samples. In particular, using noisy object detections from the unlabeled training video, we devise a loss requiring that within an individual training video, each separated audio track should be distinguishable as its proper object. For example, when two training instances both contain a guitar plus other instruments, there is pressure to make the separated guitar tracks consistently identifiable.
See Fig.~\ref{fig:concept}.

We call our idea ``co-separation" as a loose analogy to image co-segmentation~\cite{rother2006cosegmentation}, whereby jointly segmenting two related images can be easier than segmenting them separately, since it allows disentangling a shared foreground object from differently cluttered backgrounds.  Note, however, that our co-separation operates during training only; unlike co-segmentation, at test time our method performs separation on an individual video input.

Our method design offers the following advantages. First, co-separation allows training with ``in the wild" sound mixes. It has the potential to benefit from the variability and richness of unlabeled multi-source video. Second, it enhances the supervision beyond ``mix-and-separate". By enforcing separation \emph{within a single video} at the object-level, our approach exposes the learner to natural correlations between sound sources. Finally, objects with similar appearance from different videos can partner with each other to separate their sounds jointly, thereby regularizing the learning process. In this way, our method is able to learn well from multi-source videos, and it can successfully separate an object sound in a test video even if the object has never been observed individually during training.

We experiment on three benchmark datasets and demonstrate the advantages discussed above. Our approach yields state-of-the-art results on separation and denoising. Most notably, it outperforms the prior methods and  baselines by a large margin when learning from noisy AudioSet~\cite{gemmeke2017audio} videos. Overall co-separation is a promising direction to learn audio-visual separation from multi-source videos.
\section{Related Work}
\vspace{-0.05in}

\paragraph{Audio-Only Source Separation}
Audio source separation has a rich history in signal processing. While many methods assume audio captured by multiple microphones, some tackle the ``blind" separation problem with single-channel audio~\cite{hyvarinen2000independent,ellis1996prediction,virtanen2007monaural,fevotte2009nonnegative}, most recently with deep learning~\cite{huang2014deep,hershey2016deep,stoller2018adversarial}. Mix-and-separate style training is now commonly used for audio-only source separation to create artificial training examples~\cite{huang2015joint,hershey2016deep,yu2017permutation}. Our approach adapts the mix-and-separate idea. However, different from all of the above, we leverage visual object detection to guide sound source separation. Furthermore, as discussed above, our co-separation framework is more flexible in terms of training data and can generalize to multi-source videos.

\vspace{-0.2in}
\paragraph{Audio-Visual Source Separation}
Early methods for audio-visual source separation focus on mutual information~\cite{fisher2001learning}, subspace analysis~\cite{smaragdis2003audio,pu2017audio}, matrix factorization~\cite{parekh2017motion,sedighin2016two}, and correlated onsets~\cite{barzelay2007harmony,li2017see}. Recent methods leverage deep learning for separating speech~\cite{ephrat2018looking,owens2018audio,afouras2018conversation,gabbay2017visual}, musical instruments~\cite{zhao2018sound,gao2019visualsound,zhao2019som}, and other objects~\cite{gao2018objectSounds}. Similar to the audio-only methods, almost all use a ``mix-and-separate" training paradigm to perform \emph{video-level} separation by artificially mixing training videos. In contrast, we perform source separation at the \emph{object level} to explicitly model sounds coming from visual objects, and our model enforces separation \emph{within} a video during training.

Most related to our work are the ``sound of pixels" (SoP)~\cite{zhao2018sound} and multi-instance learning (AV-MIML)~\cite{gao2018objectSounds} approaches.
AV-MIML~\cite{gao2018objectSounds} also focuses on learning object sound models from unlabeled video, but its two-stage method relies on NMF to perform separation, which limits its performance and practicability. Furthermore, whereas AV-MIML simply uses image classification to obtain weak labels on video frames, our approach detects localized objects and our end-to-end network learns visual object representations in concert with the audio streams. SoP~\cite{zhao2018sound} outputs a sound for each pixel, whereas we predict sounds for visual objects with the help of a pre-trained object detector. More importantly, SoP works best when clean solo videos are available to perform video-level ``mix-and-separate" training. Our method instead disentangles mixed sounds of objects within an individual training video, allowing more flexible training with multi-source data (though unlike~\cite{zhao2018sound} we do require an object detection step).  


\vspace{-0.2in}
\paragraph{Localizing Sounds in Video Frames}
Localization entails identifying the pixels where the sound of a video  comes from, but not separating the audio~\cite{kidron2005pixels,hershey2000audio,fisher2001learning,arandjelovic2017objects,Senocak_2018_CVPR,tian2018audio}. Different from all these methods, our goal is to \emph{separate} the sounds of multiple objects from a single-channel signal.  We localize potential sound sources via object detection, and use the localized object regions to guide the separation learning process.

\paragraph{Generating Sounds from Video}
Sound generation methods synthesize a sound track from a visual input~\cite{owens2016visually,zhou2017visual,chen2017}. Given both visual input and monaural audio, recent methods generate spatial (binaural or ambisonic) audio~\cite{gao2019visualsound,morgadoNIPS18}. Unlike any of the above, our work aims to \emph{separate} an existing real audio track, not synthesize plausible new sounds.
\section{Approach}
Our approach leverages localized object detection to visually guide audio source separation. We first formalize our object-level audio-visual source separation task (Sec.~\ref{sec:formulation}). Then we introduce our framework for learning object sound models from unlabeled video and our \textsc{Co-Separation} deep network architecture (Sec.~\ref{sec:network}). Finally, we present our training criteria and inference procedures (Sec.~\ref{sec:trainingAndInference}).

\subsection{Problem Formulation}~\label{sec:formulation}
Given an unlabeled video clip $V$ with accompanying audio $x(t)$, we denote the set of $N$ objects detected in the video frames as $\mathcal{V} = \{O_1,\ldots, O_N\}$. We treat each object as a potential sound source, and $x(t) = \sum_{n=1}^{N}s_n(t)$ is the observed single-channel linear mixture of these sources, where $s_n(t)$ are time-discrete signals responsible for each object. Our goal of object-level audio-visual source separation is to separate the sound $s_n(t)$ for each object $O_n$ from $x(t)$.

Following~\cite{huang2015joint,hershey2016deep,yu2017permutation,zhao2018sound,owens2018audio,gao2019visualsound,ephrat2018looking}, we start with the commonly adopted ``mix-and-separate" idea to self-supervise source separation. Given two training videos $V_1$ and $V_2$ with corresponding audios $x_1(t)$ and $x_2(t)$, we use a pre-trained object detector to find objects in both videos. Then, we mix the audios of the two videos and obtain the mixed signal $x_m(t) = x_1(t) + x_2(t)$. The mixed audio $x_m(t)$ is transformed into a magnitude spectrogram $X^M \in \mathbb{R}_{+}^{F \times N}$ consisting of $F$ frequency bins and $N$ short-time Fourier transform (STFT)~\cite{griffin1984signal} frames, which encodes the change of a signal's frequency and phase content over time.

\begin{figure*}[t]
    \center
    \includegraphics[scale=0.83]{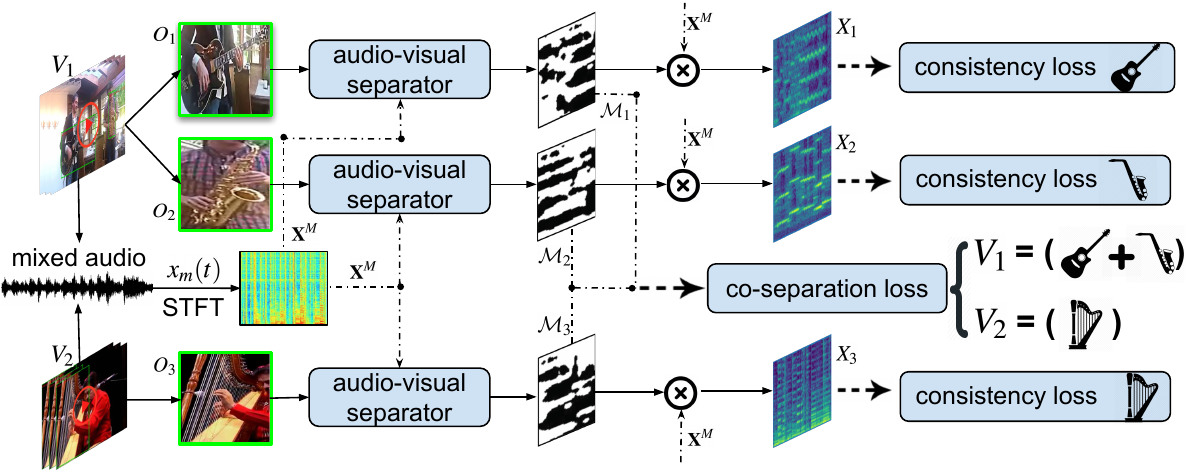}
    \caption{Co-separation training pipeline: our object-level co-separation framework first automatically detects objects in a pair of videos, then mixes the audios at the video-level, and separates the sounds for each visual object. The network is trained by minimizing the combination of the co-separation and object-consistency losses defined in Sec.~\ref{sec:network}.}
    \label{fig:network2}
    \vspace*{-0.1in}
\end{figure*}

Our learning objective is to separate the sound each object makes from $x_m(t)$ conditioned on the localized object regions. For example, Fig.~\ref{fig:network2} illustrates a scenario of mixing two videos $V_1$ and $V_2$ with two objects $O_1$, $O_2$ detected in $V_1$ and one object $O_3$ detected in $V_2$. The goal is to separate $s_1(t)$, $s_2(t)$, and $s_3(t)$ for objects $O_1$, $O_2$, and $O_3$ from the mixture signal $x_m(t)$, respectively. To perform separation, we predict a spectrogram mask $\mathcal{M}_n$ for each object. We use real-valued ratio masks and obtain the predicted magnitude spectrogram by soft masking the mixture spectrogram: $X_n = X^M \times \mathcal{M}_n$. Finally, we use the inverse short-time Fourier transform (ISTFT)~\cite{griffin1984signal} to reconstruct the waveform sound for each object source.

\begin{figure}
    \center
    \includegraphics[scale=0.53]{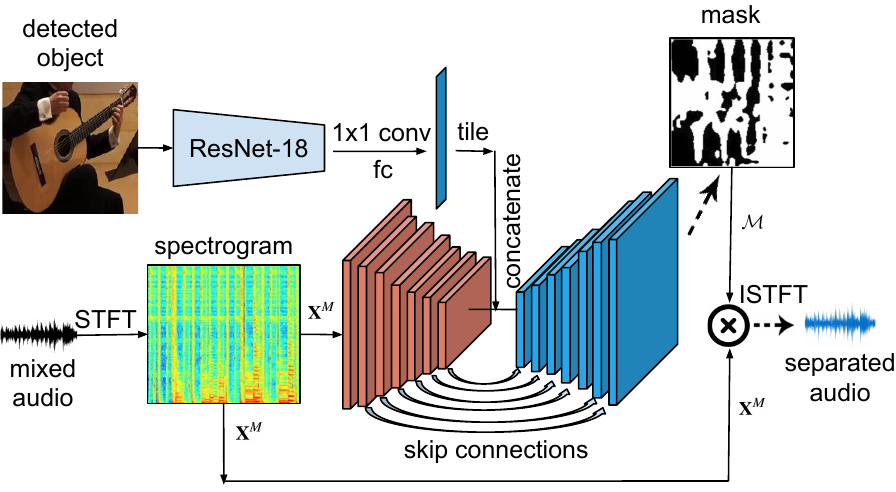}
    \caption{Our audio-visual separator network takes a mixed audio signal and a detected object from its accompanying video as input, and performs joint audio-visual analysis to separate the portion of sound responsible for the input object region.}
    \label{fig:network1}
    \vspace*{-0.1in}
\end{figure}

Going beyond video-level mix-and-separation, the key insight of our approach is to simultaneously enforce separation \emph{within a single video} at the object level. This enables our method to learn object sound models even from multi-source training videos. Our new co-separation framework can capture the correlations between sound sources and is able to learn from noisy Web videos, as detailed next.  

\subsection{Co-Separation Framework}\label{sec:network}
Next we present our \textsc{Co-Separation} training framework and our network architecture to perform separation.

\vspace*{-0.2in}
\paragraph{Object Detection}
Firstly, we train an object detector for a vocabulary of $C$ objects.  In general, this detector should cover any potential sound-making object categories that may appear in training videos.  Our implementation uses the Faster R-CNN~\cite{ren2015faster} object detector with a ResNet-101~\cite{he2016deep} backbone trained with Open Images~\cite{OpenImages2}.  For each unlabeled training video, we use the pre-trained object detector to automatically\footnote{No manual object annotations are used for co-separation train-/testing.} find objects in all video frames. Then, we gather all object detections across frames to obtain a video-level pool of objects. See Supp. for details.

\vspace*{-0.2in}
\paragraph{Audio-Visual Separator}
We use the detected object regions to guide the source separation process. Fig.~\ref{fig:network1} illustrates our audio-visual separator network that performs audio-visual feature aggregation and source separation. A related design for multi-modal feature fusion is also used in~\cite{gao2019visualsound,morgadoNIPS18,owens2018audio} for audio spatialization and separation. However, unlike those models, our separator network combines the visual features of a localized object region and the audio features of the mixed audio to predict a magnitude spectrogram mask for source separation.

The network takes a detected object region and the mixed audio signal as input, and separates the portion of the sound responsible for the object. We use a ResNet-18 network to extract visual features after the $4^{th}$ ResNet block with size $(H/32) \times (W/32) \times D$, where $H,~W,~D$ denote the frame and channel dimensions. We then pass the visual feature through a $1 \times 1$ convolution layer to reduce the channel dimension, and use a fully-connected layer to obtain an aggregated visual feature vector.

On the audio side, we adopt a U-NET~\cite{ronneberger2015u} style network for its effectiveness in dense prediction tasks, similar to~\cite{zhao2018sound,owens2018audio,gao2019visualsound}. The network takes the magnitude spectrogram $X^M$ as input and passes it through a series of convolution layers to extract an audio feature of dimension $(T/128) \times (F/128) \times D$. We replicate the visual feature vector $(T/128) \times (F/128)$ times, tile them to match the audio feature dimension, and then concatenate the audio and visual feature maps along the channel dimension. Then a series of up-convolutions are performed on the concatenated audio-visual feature map to generate a multiplicative spectrogram mask $\mathcal{M}$. We find spectrogram masks to work better than direct prediction of spectrograms or raw waveforms for source separation, confirming reports in~\cite{wang2018supervised,ephrat2018looking,gao2019visualsound}. The separated spectrogram for the input object is obtained by multiplying the mask and the spectrogram of the mixed audio:  $ X = X^M \times \mathcal{M}$. Finally, ISTFT is applied to the spectrogram to produce the separated real-time signal. 

\vspace*{-0.2in}
\paragraph{Co-Separation}
Our proposed \textsc{co-separation} framework first detects objects in a pair of videos, then mixes their audios at the video level, and finally separates the sounds for each detected object class. As shown in Fig.~\ref{fig:network2}, for each video pair, we randomly sample a high confidence object window for each class detected in either video, and use the localized object region to guide audio source separation using the audio-visual separator network. For each object $O_n$, we predict a mask $\mathcal{M}_n$, and then generate the corresponding magnitude spectrogram. 

Let $\mathcal{V}_1$ and $\mathcal{V}_2$ denote the set of objects for the two videos. We want to separate the sounds of their corresponding objects from the audio mixture of $V_1$ and $V_2$. For each video, summing up the separated sounds of all objects should ideally reconstruct the audio signal for that video. Namely, 
\vspace*{-0.05in}
\begin{equation}
	x_1(t) = \sum_{i}^{|\mathcal{V}_1|} s_i(t) \quad\text{and}\quad x_2(t) = \sum_{i}^{|\mathcal{V}_2|} s_i(t),
	\label{equation1}
\end{equation}
where $|\mathcal{V}_1|$ and $|\mathcal{V}_2|$ are the number of detected objects for $V_1$ and $V_2$. For simplicity of notation, we defer presenting how we handle background sounds (those unattributable to detected objects) until later in this section. Because we are operating in the frequency domain, the above relationship will only hold approximately due to phase interference.  As an alternative, we approximate Eq.~(\ref{equation1}) by enforcing the following relationship on the predicted magnitude spectrograms:
\begin{equation}
\vspace*{-0.05in}
	X^{V_1} \approx \sum_{i}^{|\mathcal{V}_1|} X_i \quad\text{and}\quad X^{V_2} \approx \sum_{i}^{|\mathcal{V}_2|} X_i,
\end{equation}
where $X^{V_1}$ and $X^{V_2}$ are the magnitude spectrograms for $x_1(t)$ and $x_2(t)$. Therefore, we minimize the following \emph{co-separation loss} over the separated magnitude spectrograms:
\begin{equation}
\vspace*{-0.05in}
	L_{\emph{co-separation\_spect}} = ||\sum_{i=1}^{|\mathcal{V}_1|} X_i - X^{V_1}||_1 + ||\sum_{i=1}^{|\mathcal{V}_2|} X_i - X^{V_2}||_1,
\end{equation}
which approximates to minimizing the following loss function over their predicted ratio masks:
\begin{equation}
\vspace*{-0.05in}
	L_{\emph{co-separation\_mask}} = ||\sum_{i=1}^{|\mathcal{V}_1|} \mathcal{M}_i - \mathcal{M}^{V_1}||_1 + ||\sum_{i=1}^{|\mathcal{V}_2|} \mathcal{M}_i - \mathcal{M}^{V_2}||_1,
	\label{eq:equation4}
\end{equation}
where $\mathcal{M}^{V_1}$ and $\mathcal{M}^{V_2}$ are the ground-truth spectrogram ratio masks for the two videos, respectively. Namely,
\begin{equation}
	\vspace{-0.05in}
	\mathcal{M}^{V_1} = \frac{X^{V_1}}{X^{V_1} + X^{V_2}}\quad\text{and}\quad \mathcal{M}^{V_2} = \frac{X^{V_2}}{X^{V_1} + X^{V_2}}.
\end{equation}
In practice, we find that computing the loss over masks (vs.~spectograms) makes the network easier to learn. We hypothesize that the sigmoid after the last layer of the audio-visual separator bounds the masks, making them more constrained and structured compared to spectrograms. In short, the proposed co-separation loss provides supervision to the network to only separate the audio portion responsible for the input visual object, so that the corresponding audios for each of the pair of input videos can be reconstructed.

In addition to the co-separation loss that enforces separation, we also introduce an \emph{object-consistency loss} for each predicted audio spectrogram.  The intuition is that if the sources are well-separated, the predicted ``category" of the separated spectrogram should be consistent with the category of the visual object that initially guides its separation. Specifically, for the predicted spectrogram of each object, we introduce another ResNet-18 \emph{audio} classifier\footnote{The ResNet-18 audio classifier is ImageNet pre-trained to accelerate convergence, but \emph{not} pre-trained for audio classification. Our co-separation training aims to automatically discover the audio classes.} that targets the weak labels of the input visual objects. We use the following cross-entropy loss:
\begin{equation}
	\vspace{-0.05in}
	L_{\emph{object-consistency}} = \frac{1}{|\mathcal{V}_1| + |\mathcal{V}_2|} \sum_{i=1}^{|\mathcal{V}_1| + |\mathcal{V}_2|} \sum_{c=1}^{C} -y_{i,c}\log(p_{i,c}),\label{eq:consist}
\end{equation}
where $C$ is the number of classes, $y_{i,c}$ is a binary indicator on whether $c$ is the correct class for predicted spectrogram $X_i$, and $p_{i,c}$ is the predicted probability for class $c$. We stress that these audio ``classes" are \emph{discovered} during training; we have no pre-trained sound models for different objects.

Not all sounds in a video will be attributable to a visually detected object.  To account for ambient sounds, off-screen sounds, and noise, we incorporate a $C+1^{st}$ ``adaptable" audio class, as follows. During training, we pair each video with a visual scene feature in addition to the detected objects from the pre-trained object detector. Then an additional mask $\mathcal{M}_{\text{adapt}}$ responsible for the scene context is also predicted in Eq.~(\ref{eq:equation4}) for both $V_1$ and $V_2$ to be optimized jointly. This step arms the network with the flexibility to assign noise or unrelated sounds to the ``adaptable" class, leading to cleaner separation for sounds of the detected visual objects. These adaptable objects (ideally ambient sounds, noise, \etc) are collectively designated as having the ``extra" $C+1^{st}$ audio label. The separated spectrograms for these adaptable objects are also trained to match their category label by the object-consistency loss in Eq.~(\ref{eq:consist}). 

Putting it all together, during training the network needs to discover separations for the multi-source videos that 1) minimize the co-separation loss, such that the two source videos' object sounds reassemble to produce their original video-level audio tracks, respectively, while also 2) minimizing the object consistency loss, such that separated sounds for \emph{any} instances of the same visual object are reliably identifiable as that sound.  We stress that our model achieves the latter \emph{without} any pre-trained audio model and \emph{without} any single-source audio examples for the object class.  The object consistency loss only knows that same-object sounds should be similar after training the network---not what any given object is expected to sound like.

\subsection{Training and Inference}~\label{sec:trainingAndInference}
We minimize the following combined loss function and train our network end to end:
\begin{equation}
	\vspace{-0.05in}
	L = L_{\emph{co-separation\_mask}} + \lambda L_{\emph{object-consistency}},
	\label{equation7}
\end{equation}
where $\lambda$ is the weight for the object-consistency loss. 

We use per-pixel $L1$ loss for the co-separation loss, and weight the gradients by the magnitude of the spectrogram of the mixed audio. The network uses the weighted gradients to perform back-propagation, thereby emphasizing predictions on more informative parts of the spectrogram.

During testing, our model  takes a \emph{single} realistic multi-source video to perform source separation. Similarly, we first detect objects in the video frames by using the pre-trained object detector. For each detected object class, we use the most confident object region(s) as the visual input to separate the portion of the sound responsible for this object category from its accompanying audio. We use a sliding window approach to process videos segment by segment with a small hop size, and average the audio predictions on all overlapping parts.  

We perform audio-visual source separation on video clips of 10s, and we pool all the detected objects in the video frames. Therefore, our approach assumes that each detected object within this period of 10s is a potential sound source, although it may only sound in some of the frames. For objects that are detected but do not make sound at all, we treat it as learning noise and expect our deep network to adapt by learning from large-scale training videos. We leave it as future work to explicitly model silent visual objects. 
\vspace*{-0.05in}
\section{Experiments}\label{sec:results}
\vspace*{-0.05in}
We now validate our approach for audio-visual source separation and compare to existing methods.

\subsection{Datasets}
\paragraph{MUSIC}
This MIT dataset contains YouTube videos crawled with keyword queries~\cite{zhao2018sound}. It contains 685 untrimmed videos of musical solos and duets, with 536 solo videos and 149 duet videos. The dataset is relatively clean and collected for the purpose of training audio-visual source separation models. It includes 11 instrument categories: accordion, acoustic guitar, cello, clarinet, erhu, flute, saxophone, trumpet, tuba, violin and xylophone. Following the authors' public dataset file of video IDs, we hold out the first/second video in each category as validation/test data, and the rest as training data. We split all videos into 10s clips during both training and testing, for a total of 8,928/259/269 train/val/test clips, respectively.
\vspace*{-0.15in}
\paragraph{AudioSet-Unlabeled}
AudioSet~\cite{gemmeke2017audio} consists of challenging 10s video clips, many of poor quality and containing a variety of sound sources. Following~\cite{gao2018objectSounds}, we filter the dataset to extract video clips of 15 musical instruments. We use the videos from the ``unbalanced" split for training, and videos from the ``balanced" split as validation/test data, for a total of 113,756/456/456 train/val/test clips, respectively.
\vspace*{-0.15in}
\paragraph{AudioSet-SingleSource} 
A dataset assembled by~\cite{gao2018objectSounds} of AudioSet videos containing only a single sounding object. We use the 15 videos (from the ``balanced" split) of musical instruments for evaluation only.   
\vspace*{-0.15in}
\paragraph{AV-Bench}
This dataset contains the benchmark videos (Violin Yanni, Wooden Horse, and Guitar Solo) used in previous studies~\cite{gao2018objectSounds,pu2017audio} on visually-guided audio denoising.

\vspace*{0.05in}

On both MUSIC and AudioSet, we compose the test sets following standard practice~\cite{barzelay2007harmony,zhao2018sound,owens2018audio,gao2018objectSounds}---by mixing the audio from two single-source videos.  This ensures the ground truth separated sounds are known for quantitative evaluation. There are 550 and 105 such test pairings for MUSIC and AudioSet, respectively (the result of pairwise mixing 10 random clips per the 15 classes for MUSIC and pairwise mixing all 15 clips for AudioSet).  For qualitative results (Supp.), we apply our method to real multi-source test videos. In either case, we \emph{train} our method with multi-source videos, as specified below.

\subsection{Implementation Details}\label{exp:implementaiton}
Our \textsc{Co-Separation} deep network is implemented in PyTorch. For all experiments, we sub-sample the audio at 11kHz, and the input audio sample is approximately 6s long. STFT is computed using a Hann window size of 1022 and a hop length of 256, producing a $512 \times 256$ Time-Frequency audio representation. The spectrogram is then re-sampled on a log-frequency scale to obtain a $T \times F$ magnitude spectrogram of $T = 256, F = 256$. The settings are the same as~\cite{zhao2018sound} for fair comparison. 

Our object detector is trained on images of $C=15$ object categories from the Open Images dataset~\cite{OpenImages2}. We filter out low confidence object detections for each video, and keep the top two\footnote{This is the number of objects detected in most training videos; relaxing this limit does not change the overall results (see Supp.).} detected categories. See Supp.~for details. During co-separation training, we randomly sample 64 pairs of videos for each batch. We sample a confident object detection for each class as its input visual object, paired with a random scene image sampled from the ADE dataset~\cite{zhou2017scene} as the adaptable object. The object window is resized to $256 \times 256$, and a randomly cropped $224 \times 224$ region is used as the input to the network. We use horizontal flipping, color and intensity jittering as data augmentation. $\lambda$ is set to 0.05 in Eq.~(\ref{equation7}). The network is trained using an Adam optimizer with weight decay $1 \times 10^{-4}$ with the starting learning rate set to $1 \times 10^{-4}$. We use a smaller starting learning rate of $1 \times 10^{-5}$ for the ResNet-18 visual feature extractor because it is pre-trained on ImageNet.

\begin{table*}[t]
\begin{tabular}{c?{0.5mm}ccc?{0.5mm}ccc}
\multirow{2}{*}{} & \multicolumn{3}{c?{0.5mm}}{Single-Source} & \multicolumn{3}{c}{Multi-Source} \\ \cline{2-7} 
                  & SDR    & SIR    & SAR    & SDR    & SIR    & SAR     \\ \specialrule{.12em}{.1em}{.1em}
Sound-of-Pixels~\cite{zhao2018sound}   &    7.30     &     11.9         &   \textbf{11.9}      & 6.05     &  9.81  &   \textbf{12.4}       \\ 
AV-Mix-and-Separate        &   3.16      &    6.74    &   8.89     &    3.23      &    7.01      &   9.14       \\ 
NMF-MFCC~\cite{spiertz2009source}        &   0.92      &    5.68    &     6.84   &     0.92      &    5.68    &     6.84           \\ 
\textsc{Co-Separation} 
(Ours)              &    \textbf{7.38}     &  \textbf{13.7}     &    10.8    &    \textbf{7.64}       &    \textbf{13.8}      &    11.3      \\ \specialrule{.12em}{.1em}{.1em}
\end{tabular}

\caption{Average audio source separation results on a held out MUSIC test set. We show the performance of our method and the baselines when training on only single-source videos (solo) and multi-source videos (solo + duet). NMF-MFCC is non-learned, so its results do not vary across training sets. Higher is better for all metrics.  Note that SDR and SIR capture separation accuracy; SAR captures only the absence of artifacts (and hence can be high even if separation is poor). Standard error is approximately 0.2 for all metrics.}
\label{Tab:separation1}
\vspace*{-0.1in}
\end{table*}

\subsection{Quantitative Results on Source Separation}
\vspace*{-0.05in}
\begin{table}[t]
\begin{tabular}{c?{0.5mm}ccc}
                  & SDR    & SIR    & SAR  \\ \specialrule{.12em}{.1em}{.1em}
Sound-of-Pixels~\cite{zhao2018sound}   &  1.66 &  3.58 &   11.5   \\ 
AV-MIML~\cite{gao2018objectSounds}    & 1.83 & -  & -   \\ 
AV-Mix-and-Separate             &  1.68& 3.30 & 12.2  \\ 
NMF-MFCC~\cite{spiertz2009source}    & 0.25 & 4.19  & 5.78   \\ 
\textsc{Co-Separation} (Ours)               & \textbf{4.26}  & \textbf{7.07}  &  \textbf{13.0}    \\ \specialrule{.12em}{.1em}{.1em}
\end{tabular}
\caption{Average separation results on AudioSet test set. Standard error is approximately 0.3.}
\label{Tab:separation2}
\vspace*{-0.1in}
\end{table}
We compare to the following baselines:
\vspace*{-0.05in}
\begin{itemize}
\itemsep0em
\item \textbf{Sound-of-Pixels~\cite{zhao2018sound}:} We use the authors' publicly available code\footnote{\url{https://github.com/hangzhaomit/Sound-of-Pixels}} to train 1-frame based models with ratio masks for fair comparison. Default settings are used for other hyperparameters. 
\item \textbf{AV-Mix-and-Separate:} A ``mix-and-separate" baseline using the same audio-visual separation network as our model to do \emph{video}-level separation.  We use multi-label hinge loss to enforce video-level consistency, i.e., the class of each separated spectrogram should agree with the objects present in that training video.
\item \textbf{AV-MIML~\cite{gao2018objectSounds}:} An existing audio-visual source separation method that uses audio bases learned from unlabeled videos to supervise an NMF separation process. The audio bases are learned from a deep multi-instance multi-label (MIML) learning network. We use the results reported in~\cite{gao2018objectSounds} for AudioSet and AV-Bench; the authors do not report results in SDR and do not report results for MUSIC.
\item \textbf{NMF-MFCC~\cite{spiertz2009source}:} An off-the-shelf audio-only method that performs NMF based source separation using Mel frequency cepstrum coefficients (MFCC). This non-learned baseline is a good representation of a well established pipeline for audio-only source separation~\cite{virtanen2003sound,innami2012nmf,jaiswal2011clustering,guo2015nmf}.
\item \textbf{AV-Loc~\cite{pu2017audio}, JIVE~\cite{lock2013joint}, Sparse CCA~\cite{kidron2005pixels}:}  We use results reported in~\cite{gao2018objectSounds} to compare to these methods for the audio denoising benchmark AV-Bench.
\end{itemize}
\vspace*{-0.05in}

We use the widely used mir eval library~\cite{raffel2014mir_eval} to evaluate the source separation and report the standard metrics: Signal-to-Distortion Ration (SDR), Signal-to-Interference Ratio (SIR), and Signal-to-Artifact Ratio (SAR).

\vspace*{-0.15in}
\paragraph{Separation Results.} Tables~\ref{Tab:separation1} and~\ref{Tab:separation2} show the results for the MUSIC and AudioSet datasets, respectively.

Table~\ref{Tab:separation1} presents results on MUSIC as a function of the training source: single-source videos (solo) or multi-source videos (solo + duet). Our method consistently outperforms all baselines in separation accuracy, as captured by the SDR and SIR metrics.\footnote{Note that SAR measures the \emph{artifacts} present in the separated signal, but not the separation accuracy. So, a less well-separated signal can achieve high(er) SAR values. In fact, naively copying the original input twice (i.e., doing no separation at all) results in SAR $\approx$ 80 in our setting.} While the SoP method~\cite{zhao2018sound} works well when training only on solo videos, it fails to make use of the additional duets, and its performance degrades when training on the multi-source videos. In contrast, our method actually \emph{improves} when trained on a combination of solos and duets, achieving its best performance. This experiment highlights precisely the limitation of the mix-and-separate training paradigm when presented with multi-source training videos, and it demonstrates that our co-separation idea can successfully overcome that limitation.

Our method also outperforms all baselines, including \cite{zhao2018sound}, when training on solos. Our better accuracy versus the AV-Mix-and-Separate baseline and \cite{zhao2018sound} shows that our object-level co-separation idea is essential. The NMF-MFCC baseline can only return ungrounded separated signals. Therefore, we evaluate both possible matchings and take its best results (to the baseline's advantage). Also, our gains are similar even if we give~\cite{zhao2018sound} the advantage of temporal pooling over 3 frames. Overall our method achieves large gains, and also has the benefit of matching the separated sounds to semantically meaningful visual objects in the video.

Table~\ref{Tab:separation2} shows the results when training on AudioSet-Unlabeled and testing on mixes of AudioSet-SingleSource. Our method outperforms all prior methods and the baselines by a large margin on this challenging dataset. It demonstrates that our framework can better learn from the noisy and less curated ``in the wild" videos of AudioSet, which contains many multi-source videos. See Supp.~for additional results on removing the limit of two objects per video.

Next we devise an experiment to test explicitly how well our method can learn to separate sound for objects it has not observed \emph{individually} during training. We train our model and the best baseline~\cite{zhao2018sound} on the following four categories: violin solo, saxophone solo, violin+guitar duet, and violin+saxophone duet, and test by randomly mixing and separating violin, saxophone, and guitar test solo clips. Table~\ref{Tab:separation3} shows the results. We can see that although our system is not trained on any guitar solos, it can learn better from multi-source videos that contain guitar and other sounds. Our method consistently performs well on all three combinations, while~\cite{zhao2018sound} performs well only on the violin+guitar mixture. We hypothesize the reason is that it can learn by mixing the large quantity of violin solos and the guitar solo moments \emph{within} the duets to perform separation, but it fails to disentangle other sound source correlations. Our method scores worse in terms of SAR, which again measures artifacts, but not separation quality. 

See Supp.~for additional experiments where we train \emph{only} on duets as well as an ablation study to isolate the impact of each loss term.

\begin{figure}[t]
	\centering
	\includegraphics[scale=0.26]{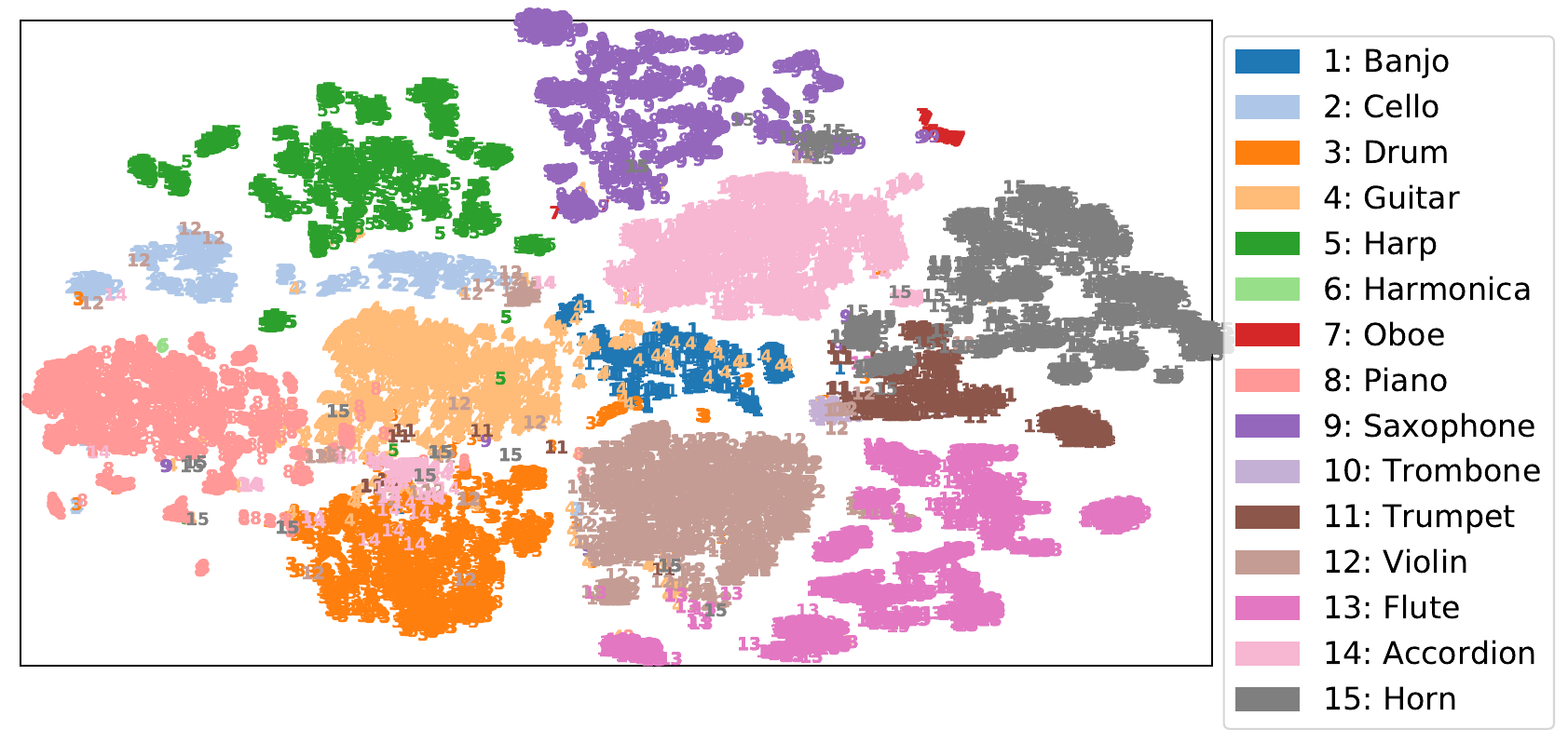}
		\includegraphics[scale=0.11]{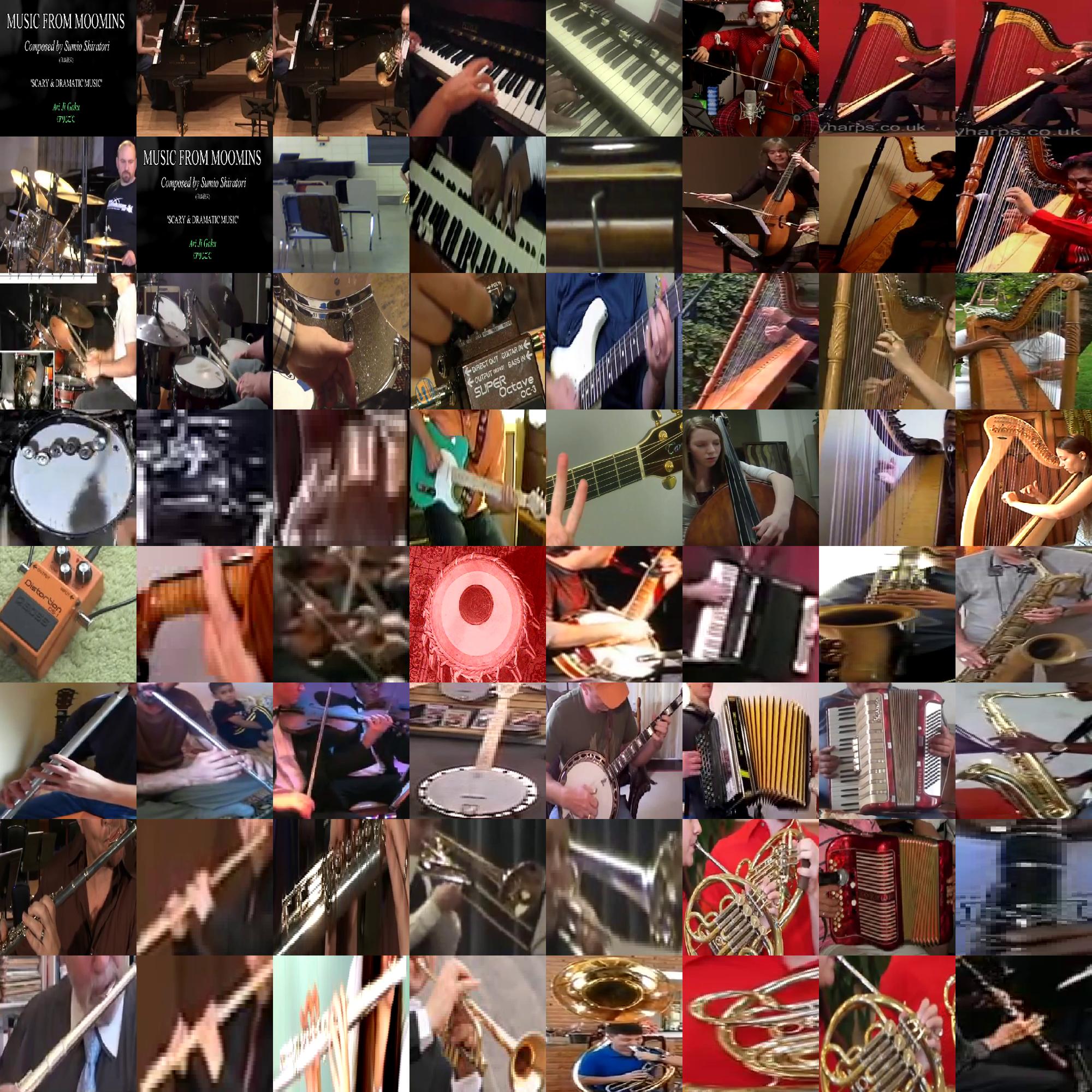}
	\caption{Embedding of separated sounds in AudioSet visualized with t-SNE in two ways: (top) categories are color-coded, and (bottom) visual objects are shown at their sound's embedding.}
	\label{fig:tsne}
	\vspace*{-0.05in}
\end{figure}

\begin{table}
\fontsize{7.8}{9.6} \selectfont
\begin{tabular}{c?{0.5mm}ccc?{0.5mm}ccc}
 & \multicolumn{3}{c?{0.5mm}}{Sound-of-Pixels~\cite{zhao2018sound}} & \multicolumn{3}{c}{\textsc{Co-Separation} (Ours)} \\ \cline{2-7} 
 & SDR   & SIR   & SAR   & SDR   & SIR   & SAR   \\ \specialrule{.12em}{.1em}{.1em}
 Violin/Saxophone &    1.52   &    1.48   &     12.9  &   8.10    &    11.7   &  11.2     \\ 
 Violin/Guitar &   6.95    &   11.2    &  15.8     &   10.6    &    16.7   &   12.3    \\ 
 Saxophone/Guitar &   0.57    &  0.90     &   16.5    &  5.08     &    7.90   &   9.34    \\ \specialrule{.12em}{.1em}{.1em}
\end{tabular}
\caption{Toy experiment to demonstrate learning to separate sounds for objects never heard individually during training.}
\label{Tab:separation3}
\vspace*{-0.1in}
\end{table}

\vspace*{-0.2in}
\paragraph{Denoising Results.}
As a side product of our audio-visual source separation system, we can also use our model to perform visually-guided audio denoising. As mentioned in Sec.~\ref{sec:trainingAndInference}, we use an additional scene image to capture ambient/unseen sounds and noise, \etc. Therefore, given a test video with noise, we can use the top detected visual object in the video to guide our system to separate out the noise.

Table~\ref{Tab:denoising} shows the results on  AV-Bench~\cite{pu2017audio,gao2018objectSounds}. Though our method learns only from unlabeled video and does not explicitly model the low-rank nature of noise as in~\cite{pu2017audio}, we obtain state-of-the-art performance on 2 of the 3 videos. The method of \cite{pu2017audio} uses motion in manually segmented regions, which may help on Guitar Solo, where the hand's motion strongly correlates with the sound.

\begin{table}[t]
\centering
\fontsize{8.5}{10.5} \selectfont
\begin{tabular}{c?{0.5mm}c|c|c}
           & Wooden Horse & Violin Yanni & Guitar Solo \\ \specialrule{.12em}{.1em}{.1em}
Sparse CCA~\cite{kidron2005pixels} &     4.36         &       5.30       &   5.71   \\ 
JIVE~~\cite{lock2013joint}      &       4.54       &              4.43 &   2.64    \\ 
AV-Loc~\cite{pu2017audio}  &     8.82         &              5.90 &   \textbf{14.1}   \\
AV-MIML~\cite{gao2018objectSounds}       &      12.3        &      7.88        &   11.4     \\ \specialrule{.12em}{.1em}{.1em}
Ours      &      \textbf{14.5}        &      \textbf{8.53}        &   11.9     \\ 
\end{tabular}
\caption{Visually-assisted audio denoising on AV-Bench, in terms of NSDR (in dB, higher is better).}
\label{Tab:denoising}
\vspace*{-0.1in}
\end{table}

\vspace*{-0.2in}
\subsection{Qualitative Results}
\paragraph{Audio-Visual Separation Video Examples.}
Our video results (see Supp.) show qualitative separation results. We use our system to discover and separate object sounds for realistic multi-source videos. They lack ground truth, but the results can be manually inspected for quality.

\vspace*{-0.15in}
\paragraph{Learned Audio Embedding.}
To \emph{visualize} that our \textsc{co-separation} network has indeed learned to separate sounds of visual objects, Fig.~\ref{fig:tsne} displays a t-SNE~\cite{maaten2008visualizing} embedding of the discovered sounds for various input objects in 20K AudioSet clips. We use the features extracted at the last layer of the ResNet-18 audio classifier as the audio representation for the separated spectrograms. The sounds our method learned from multi-source videos tend to cluster by object category, demonstrating that the separator discovers sounds characteristic of the corresponding objects.

\vspace*{-0.15in}
\paragraph{Using Discovered Sounds to Detect Objects.} Finally, we use our trained audio-visual source separation network for \emph{visual object discovery} using 912 noisy unseen videos from AudioSet. Given the pool of videos, we generate object region proposals using Selective Search~\cite{uijlings2013selective}. Then we pass these region proposals to our network together with the audio of its accompanying video, and retrieve the \emph{visual} proposals that achieve the highest \emph{audio} classification scores according to our object consistency loss.

\vspace*{0.05in}

\begin{figure}[t]
	\centering
	\includegraphics[scale=0.47]{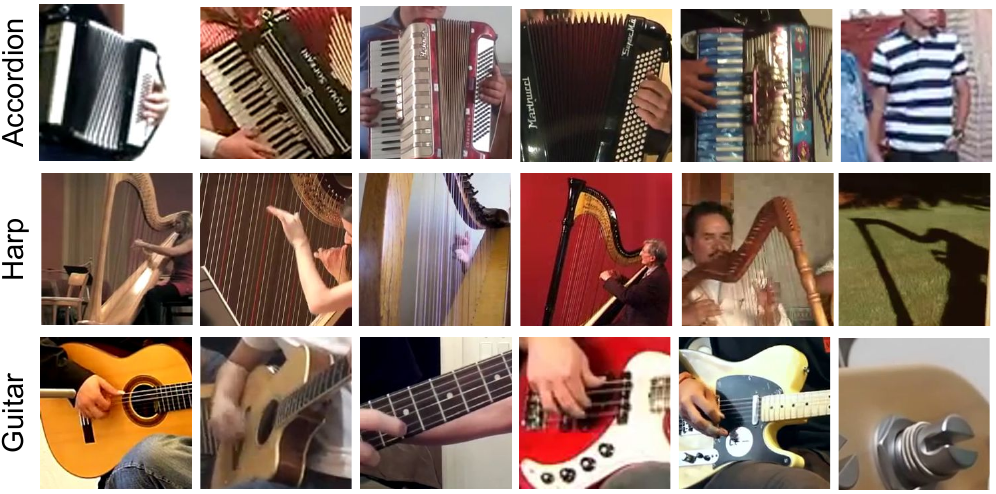}
	\caption{Top object proposals according to our discovered \emph{audio} classifier. Last column shows typical failure cases.}
	\label{fig:objectDiscovery}
	\vspace*{-0.1in}
\end{figure}

Fig.~\ref{fig:objectDiscovery} shows the top retrieved proposals for several categories after removing duplicates from the same video. We can see that our method has learned a good mapping between the visual and audio modalities; the best visual object proposals usually best activate the audio classifier. The last column shows failure cases where the wrong object is detected with high confidence. They usually come from objects of similar texture or shape, like the stripes on the man's t-shirt and the shadow of the harp.

\vspace*{-0.05in}
\section{Conclusion}

 We presented an object-level audio-visual source separation framework that associates localized object regions in videos to their characteristic sounds. Our \textsc{Co-Separation} approach can leverage noisy object detections as supervision to learn from large-scale unlabeled videos. We achieve state-of-the-art results on visually-guided audio source separation and audio denoising. As future work, we plan to explore spatio-temporal object proposals and incorporate object motion to guide separation, which may especially benefit object sounds with similar frequencies.
 

\vspace*{-0.1in}
\footnotesize
\paragraph{Acknowledgements:} Thanks to Dongguang You for help with experiments setup, and Yu-Chuan Su, Tushar Nagarajan, Santhosh Ramakrishnan and Xingyi Zhou for helpful discussions and reading paper drafts. UT Austin is supported in part by DARPA Lifelong Learning Machines.

{\small
\bibliographystyle{ieee_fullname}
\bibliography{ref_RG.bib}
}

\clearpage

\end{document}